\PassOptionsToPackage{table}{xcolor}
\documentclass{article} 
\usepackage{iclr2027_conference,times}

\usepackage{hyperref}
\usepackage{url}

\usepackage{tabularx}
\usepackage{booktabs}
\usepackage{multirow}
\usepackage{graphicx}
\usepackage{float}
\usepackage{amsmath}
\usepackage{amssymb}
\usepackage{xcolor}
\usepackage{caption}
\usepackage{subcaption}
\usepackage{algorithm}
\usepackage{listings}
\usepackage{tikz}
\usetikzlibrary{arrows.meta,calc,positioning}

\title{Few-Step Generation via Data-Space Iteration}

\author{Shanchuan Lin$^\dagger$ \quad Yansong Peng \quad Fu-Yun Wang \quad Haoqi Fan \\
ByteDance Seed \\
\texttt{peterlin@bytedance.com}$^\dagger$
}

\iclrfinalcopy 
\preprint 
\begin{document}

\maketitle

\begin{abstract}
Flow matching has emerged as a scalable paradigm for training high-quality generative models, but sampling from the learned probability flow requires many network evaluations. Distillation can reduce this cost to one or a few evaluations; however, one-step generation often sacrifices quality, making few-step generation the practical operating regime. Existing few-step methods perform their iterative computation along the probability flow and therefore require a fixed, manually chosen timestep discretization. This discretization is often chosen heuristically and is expensive to tune; it may also be restrictive when refinement difficulty differs across samples or spatial locations. We introduce data-space iteration, a few-step generation framework that removes flow discretization altogether. Starting from noise, a shared generator directly refines its prediction in data space, with every iteration trained to produce the best sample permitted by its capacity. Our formulation integrates with distribution matching distillation (DMD) with minimal changes, enabling a controlled comparison between iteration methods under matched training settings. On class-conditional ImageNet $256\times256$, data-space iteration outperforms standard discretization baselines and matches or improves upon variants selected through schedule search, without requiring schedule-specific training. These results show that data-space iteration provides a simple and effective alternative to discretized flow-space iteration for fast generation.
\end{abstract}

\begin{figure}[H]
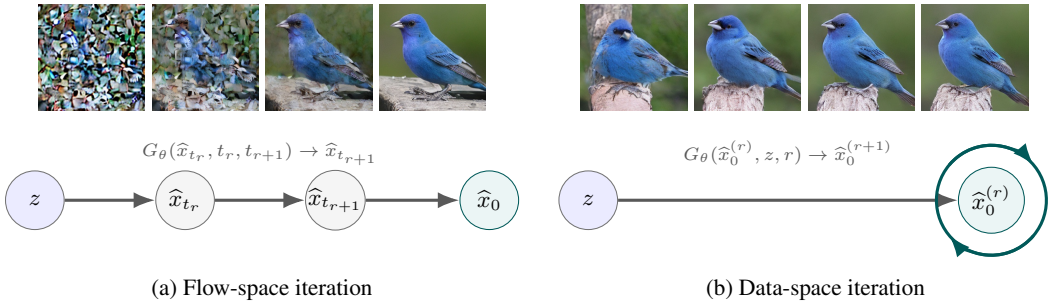

    \centering
    \resizebox{\linewidth}{!}{%
    \begin{tikzpicture}[
        font=\small,
        >=Latex,
        state/.style={
            circle,
            draw=black!55,
            fill=black!4,
            minimum size=8mm,
            inner sep=0pt,
            align=center
        },
        datastate/.style={
            circle,
            draw=teal!70!black,
            fill=teal!8,
            minimum size=9mm,
            inner sep=0pt,
            align=center
        },
        flowarrow/.style={->, very thick, black!65},
        refinearrow/.style={->, very thick, teal!70!black}
    ]
        \foreach \i/\x in {1/0.775,2/2.325,3/3.875,4/5.425} {
            \node[inner sep=0pt, outer sep=0pt] (flowimg\i) at (\x,1.95)
                {\includegraphics[width=1.45cm]{img/hero/flow_\i.jpg}};
        }

        \foreach \i/\x in {1/8.175,2/9.725,3/11.275,4/12.825} {
            \node[inner sep=0pt, outer sep=0pt] (dataimg\i) at (\x,1.95)
                {\includegraphics[width=1.45cm]{img/hero/data_\i.jpg}};
        }

        \node[state, fill=blue!8] (zflow) at (0,0) {$z$};
        \node[state] (xt) at (2.05,0) {$\widehat{x}_{t_r}$};
        \node[state] (xs) at (4.1,0) {$\widehat{x}_{t_{r+1}}$};
        \node[state, draw=teal!70!black, fill=teal!8] (x0) at (6.2,0) {$\widehat{x}_0$};

        \draw[flowarrow] (zflow) -- (xt);
        \draw[flowarrow] (xt) -- node[above=12pt, font=\scriptsize, fill=white, inner sep=1pt]
            {$G_\theta(\widehat{x}_{t_r},t_r,t_{r+1})\to\widehat{x}_{t_{r+1}}$} (xs);
        \draw[flowarrow] (xs) -- (x0);

        \node at (3.1,-1.22)
            {(a) Flow-space iteration};

        \node[state, fill=blue!8] (zdata) at (7.55,0) {$z$};
        \node[datastate] (xr) at (13.05,0) {$\widehat{x}_0^{(r)}$};
        \draw[flowarrow] (zdata) -- node[above=12pt, font=\scriptsize, fill=white, inner sep=1pt]
            {$G_\theta(\widehat{x}_0^{(r)},z,r)\to\widehat{x}_0^{(r+1)}$} (xr);
        \draw[refinearrow]
            ($(xr)+(140:0.75cm)$)
            arc[start angle=140, end angle=-140, radius=0.75cm];
        \draw[refinearrow]
            ($(xr)+(320:0.75cm)$)
            arc[start angle=320, end angle=40, radius=0.75cm];
        \node at (10.65,-1.22)
            {(b) Data-space iteration};
    \end{tikzpicture}
    }
    \caption{
        (a) Existing few-step generators perform iterative sampling in probability-flow space,
        requiring a predefined timestep discretization that can be suboptimal.
        (b) Our method carries out all refinements in data space,
        removing the discretization bottleneck and yielding better performance.
    }
    \label{fig:hero}
\end{figure}

\section{Introduction}

Flow matching~\citep{flowmatching} has become a scalable pretraining paradigm for modern generative models. By learning a continuous probability flow between simple noise and the data distribution, it provides a stable and flexible foundation for high-capacity image and video generators~\citep{seaweed,seedance15,seedance2}. However, this strength comes with an inference-time cost: producing a high-quality sample requires numerically integrating the learned flow, often with many network evaluations. This cost has motivated a broad line of distillation methods that convert a pretrained flow-matching model into a generator that can sample in one or only a few steps~\citep{progressive,cm,shortcut,meanflow}.

In principle, distillation can aim for one-step generation. In practice, however, a single network evaluation provides limited effective computational depth, and aggressive one-step distillation often leads to visible quality degradation~\citep{apt}. Few-step generation has therefore become the more practical operating point, trading a small number of additional evaluations for substantially improved sample quality. This is also the regime most relevant to deployed high-resolution generative systems, where four-step sampling is commonly used as a favorable balance between latency and fidelity~\citep{rcm}.

Despite this progress, existing few-step generators still perform their iterative computation in probability-flow space. Whether the method denoises and then renoises to the next noise level, or directly learns maps between intermediate flow marginals, generation is organized around a manually chosen discretization of the flow-time interval. This discretization is usually treated as a fixed design choice rather than a learned component of the generator.

The fixed discretization is especially limiting for modern distillation procedures based on self-rollout backward distillation~\citep{dmd2,imagineflash}. In these methods, each generator update is trained on states produced by the generator's own preceding rollout steps. The rollout schedule must therefore be defined during training, not merely selected at inference time. In applications such as causal streaming video generation, self-rollout is even more central because the model must learn to refine its own generated history~\citep{apt2,selfforcing,causalrcm}. Yet the timestep discretization used for rollout is typically chosen heuristically~\citep{simplediffusion,sd3}. Searching over alternative schedules is expensive because each candidate schedule can require training a different distilled generator.

One intuition motivating our formulation is that there may not be a universal best timestep discretization. The difficulty of a denoising transition could depend on the sample being generated: different prompts or classes can induce samples with different semantic and spectral complexity and might therefore favor different effective step sizes. Likewise, spatial regions with different frequency content might benefit from different forms of refinement. A single global flow timestep nevertheless advances every sample and location through the same denoising schedule. This observation motivates investigating whether refinement can be organized without a fixed probability-flow discretization.

We propose \emph{data-space iteration} for few-step generation. Instead of repeatedly moving along a discretized probability-flow path, our generator performs all refinement directly in data space. Starting from an initial noise sample, the generator repeatedly updates the current data-space prediction. Every refinement step is trained to output the best sample it can, given the current estimate and the original noise. This formulation removes the timestep-discretization hyperparameter from the generator's iterative sampling process and lets the model spend all of its computation on directly improving the sample.

Our method can be trained with almost no modification to the widely used distribution matching distillation (DMD) framework~\citep{dmd1,dmd2}. The real and fake flow models, the DMD objective, and the self-rollout training structure are retained; only the generator rollout is changed from probability-flow iteration to data-space iteration. This makes the comparison especially controlled: the training recipe is held fixed while the iterative sampling space is changed.

We evaluate on class-conditional ImageNet $256\times256$ generation~\citep{imagenet}, using a pretrained SiT-XL/2~\citep{sit} flow-matching teacher backbone. Across matched DMD training settings, data-space iteration outperforms standard discretization baselines and matches or improves upon variants selected through schedule search. These results suggest that removing manually specified flow-time discretization can retain or improve generation quality while eliminating the need to train schedule-specific generators.

\section{Method}

\subsection{Flow matching background}

Let $x_0\sim p_{\mathrm{data}}$ be a data sample and $x_1\sim p_{\mathrm{noise}}$ be an independent noise sample. Flow matching connects the two distributions through a continuous flow. A common choice is linear interpolation,
\begin{equation}
    x_t = (1-t) x_0 + t x_1,
    \qquad t\in[0,1].
\end{equation}

A neural network $v_\phi$ is trained to predict the velocity of this flow,
\begin{equation}
    \frac{d x_t}{dt}=v_\phi(x_t,t).
\end{equation}

To generate a sample, we draw $x_1\sim p_{\mathrm{noise}}$ and solve this ordinary differential equation (ODE) backward from $t=1$ to $t=0$. An accurate numerical solution often requires tens or hundreds of network evaluations. Distillation replaces this costly process with a generator that produces a sample in only a few evaluations.

\subsection{Existing few-step generation paradigms}

Existing few-step methods perform iterations in the probability-flow space. We group them into two paradigms.

\paragraph{Flow-renoise iteration.}
Given a noisy state at schedule point $t_r$, the generator predicts the data endpoint,
\begin{equation}
    \widehat{x}_0=G_\theta(x_{t_r},t_r).
\end{equation}
To continue iteration, this prediction is mixed with noise at the next schedule point $t_{r+1}<t_r$,
\begin{equation}
    \widehat{x}_{t_{r+1}}=(1-t_{r+1})\widehat{x}_0+t_{r+1}\varepsilon,
    \qquad \varepsilon\sim p_{\mathrm{noise}}.
\end{equation}
The resulting noisy state is passed back to the generator, and the predict--renoise process is repeated. The original DMD method~\citep{dmd1,dmd2}, consistency models~\citep{cm,icm,scm,lcm,animatelcm}, and various other methods~\citep{apt,hypersd,add,ladd} are representatives of this paradigm.

\paragraph{Flow-map iteration.}
Instead of predicting the data endpoint at every step, flow-map methods learn a direct transition between consecutive schedule points,
\begin{equation}
    \widehat{x}_{t_{r+1}}
    =G_\theta(x_{t_r},t_r,t_{r+1}).
\end{equation}
Generation composes these maps over consecutive time intervals until it reaches $t_R=0$. Phased DMD~\citep{phaseddmd}, MeanFlow~\citep{meanflow,imeanflow,tvm,alphaflow,facm}, and many other methods~\citep{progressive,pcm,perflow,sdxllightning,animatedifflightning,afm,shortcut} are examples of this paradigm.

\paragraph{Dependence on flow-time discretization.} Existing methods use flow time to define progress during generation. They must therefore choose a schedule such as $1=t_0>t_1>\cdots>t_R=0$. For self-rollout distillation, this schedule determines the states seen during training and must be fixed before training begins. Testing another schedule generally requires training another model. A global schedule is also shared by all samples and spatial locations, which motivates considering a schedule-free alternative. Data-space iteration removes the flow-time schedule by passing a data prediction, rather than a flow state, between steps.

\subsection{Few-step generation via data-space iteration}

Given noise $z\sim p_{\mathrm{noise}}$, we initialize a data-space estimate with zeros,
\begin{equation}
    \widehat{x}_0^{(0)}=\mathbf{0},
\end{equation}
and refine it by repeatedly applying a shared generator,
\begin{equation}
    \widehat{x}_0^{(r+1)}
    =G_\theta\!\left(\widehat{x}_0^{(r)},z,r\right),
    \qquad r=0,\ldots,R-1.
\end{equation}
Here, $r$ is the refinement index and $R$ is the maximum number of network evaluations. The same source noise $z$ is provided at every step, while conditioning on $r$ allows the shared generator to use a different update at each stage.

Every generator output is a prediction in data space. Unlike flow-renoise, our method does not add noise between predictions. Unlike flow-map, it does not construct states at intermediate flow times. Each step is trained to produce a valid sample and can use the previous prediction to improve its output. Sampling therefore does not require a flow-time schedule.

\subsection{Training via distribution matching distillation}

Data-space iteration requires only a small change to standard DMD training. We retain the frozen pretrained flow model $v_{\mathrm{real}}=v_\phi$ and an auxiliary fake flow model $v_{\mathrm{fake}}=v_\psi$. The fake model is initialized from the pretrained model, $\psi\leftarrow\phi$, and is additionally conditioned on the refinement index: $v_\mathrm{fake}(x_\tau,\tau,r)$. It is trained to track the generator distribution at each refinement step.

\paragraph{Generator update.}
We sample a refinement index $r\in\{0,\ldots,R-1\}$ and noise $z\sim p_{\mathrm{noise}}$. We first run $r$ refinement steps without tracking gradients to obtain $\widehat{x}_0^{(r)}$. We then take one step with gradients,
\begin{equation}
    \widehat{x}_0^{(r+1)}
    =G_\theta\!\left(
        \widehat{x}_0^{(r)},z,r
      \right).
\end{equation}

Next, we sample an independent noise sample $\varepsilon\sim p_{\mathrm{noise}}$ and a DMD noising time $\tau$, and add noise to the new prediction,
\begin{equation}
    \widehat{x}_\tau^{(r+1)}
    =(1-\tau)\widehat{x}_0^{(r+1)}+\tau\varepsilon.
\end{equation}

The generator is trained with the DMD loss
\begin{equation}
    \mathcal{L}_\mathrm{dmd}(\theta)
    =\mathbb{E}_{z,r,\tau,\varepsilon}\!\left[
        w(\tau)
        \operatorname{sg}\!\left[
            v_{\mathrm{real}}\!\left(\widehat{x}_\tau^{(r+1)},\tau\right)
            -v_{\mathrm{fake}}\!\left(\widehat{x}_\tau^{(r+1)},\tau,r\right)
        \right]^{\!\top}\widehat{x}_0^{(r+1)}
    \right],
\end{equation}
where $\operatorname{sg}[\cdot]$ denotes stop-gradient and $w(\tau)$ is a weighting function. The difference between the real and fake flow predictions therefore provides a fixed update direction that moves the generator distribution toward the data distribution.

\paragraph{Fake-model update.}
Alternating with generator updates, we train the fake flow model on noisy generator samples. For the linear path above, its target velocity is $\varepsilon-\widehat{x}_0^{(r+1)}$, giving
\begin{equation}
    \mathcal{L}_{\mathrm{fake}}(\psi)
    =\mathbb{E}_{z,r,\tau,\varepsilon}\!\left[
        \left\|
            v_\mathrm{fake}\!\left(\widehat{x}_\tau^{(r+1)},\tau,r\right)
            -\left(\varepsilon-\widehat{x}_0^{(r+1)}\right)
        \right\|_2^2
      \right].
\end{equation}

Note that this self-rollout strategy is already standard in DMD-based few-step generators. We change only the rollout process: existing methods iterate along the probability flow, whereas our method iterates in data space. The core DMD formulation remains unchanged, so our method adds no additional computation or memory relative to a matched DMD baseline.

\paragraph{Scope of the training objective.}
Some prior DMD-based methods add auxiliary adversarial (GAN) losses~\citep{gan} or rectified-flow objectives~\citep{rectifiedflow}. We omit these components because they introduce extra loss weights and design choices, expanding the search space and making it harder to isolate the effect of the iteration method. Our goal is not to combine every available technique in pursuit of state-of-the-art performance, but to provide a simple, controlled, and reproducible setting for comparing iterative sampling methods.

\subsection{Generator implementation}

We build the generator by minimally modifying the pretrained velocity network $v_\phi(x_t,t)$. Let $g_\theta$ denote the modified network, initialized with the pretrained parameters $\theta \leftarrow \phi$. The source noise $z$ uses the original input projection. We add a separate projection for the current estimate $\widehat{x}_0^{(r)}$ and sum the two projected inputs. We also add an embedding for $r$ to the existing timestep-conditioning pathway. The original flow-time input is fixed to $t=1$ for every generator evaluation.

The generator output is
\begin{equation}
    \widehat{x}_0^{(r+1)}
    =G_\theta\!\left(\widehat{x}_0^{(r)},z,r\right)
    :=z-g_\theta\!\left(\widehat{x}_0^{(r)},z,r,t=1\right).
\end{equation}
For the linear path, the target velocity is $x_1-x_0$. At the noise endpoint, setting $x_1=z$ gives the corresponding data prediction $x_0=z-v$. This motivates subtracting the predicted velocity from $z$. The construction preserves the pretrained parameterization while adding only the inputs needed for data-space refinement.

The added parameters are often negligible, and the timestep projection MLP can be replaced with a constant $t=1$ embedding, so the size of the model actually decreases, as detailed in the appendix.

\subsection{Connection to fixed-point models}
\label{sec:fixed-point-models}

Repeatedly applying a shared generator resembles a fixed-point model~\citep{deq}, whose output is defined by an equilibrium
\begin{equation}
    x^\ast = G_\theta(x^\ast,z).
\end{equation}
Our formulation differs in two ways. First, our generator is conditioned on $r$, so its update can change across refinement steps. Second, we apply the distribution-matching objective at every step rather than only after convergence. Thus, every prefix of the rollout, up to the trained maximum $R$, is encouraged to produce a valid sample.

We also do not require the iteration to converge to a fixed point. Matching the marginal output distribution at every step does not force different steps to implement the same mapping from noise to data. To test whether a more consistent mapping is beneficial, we consider adding an optional optimal transport regularizer~\citep{afm,dmd-ot}
\begin{equation}
    \mathcal{L}_\mathrm{ot}(\theta)
    = \lambda_\mathrm{ot}\,
    \mathbb{E}_{z,r}\!\left[
        \left\|
            G_\theta\!\left(\widehat{x}_0^{(r)},z,r\right)-z
        \right\|_2^2
    \right].
\end{equation}
Combined with distribution matching, this regularization objective biases each step toward learning the same Wasserstein-2 optimal transport of $z$ to the data distribution, which may encourage successive predictions to stabilize. However, as shown in our later experiments, it does not lead to any improvement.

\section{Experiments}

\subsection{Experimental setup}

Following prior work on few-step generation, we evaluate on the ImageNet $256\times256$ class-conditional generation benchmark in the latent space~\citep{ldm}\footnote{\url{https://huggingface.co/stabilityai/sd-vae-ft-mse}}. All methods use the transformer architecture, specifically DiT-XL/2~\citep{dit}. We report Fr\'echet Inception Distance (FID)~\citep{fid} over 50k class-balanced generated samples against the ImageNet training set, where lower values indicate better quality. We measure sampling cost by the number of function evaluations (NFE). For the pretrained flow-matching backbone, we directly adopt the SiT-XL/2 checkpoint~\citep{sit}.

For DMD training, we set $R=4$ so that the trained generator supports generation with up to four steps; as shown in our later experiments, using additional steps does not further improve FID. We optimize with Adam~\citep{adam}, using a learning rate of $1\times10^{-4}$ and a batch size of $256$ throughout all experiments. We perform four optimization updates of $v_{\mathrm{fake}}$ for each update of $G_\theta$. All models are trained for $40$ epochs.

\subsection{Ablation study on data-space iteration generation}

\paragraph{DMD weighting.}
As shown in Table~\ref{tab:dmd_weighting}, the DMD timestep weighting can influence the resulting FID. We first search over loss-weighting functions $w(\tau)$ under uniform timestep sampling and find that performance is best when training places greater emphasis on high-noise timesteps. We then replace loss weighting with importance sampling over $\tau$. Under equivalent settings, importance sampling consistently outperforms loss weighting. We attribute this improvement to its more efficient use of each minibatch: rather than assigning low weights to part of a uniformly sampled batch, it allocates more samples to the timesteps that contribute most to the objective. Importance sampling also aligns the timestep distribution used to train $G_\theta$ with that used to train $v_{\mathrm{fake}}$.

\begin{table}[t]
    \centering
    \caption{Ablation studies of the data-space iteration generator. FID lower is better.}
    \label{tab:data_space_ablation}
    \begin{subtable}[t]{\textwidth}
        \centering
\caption{DMD timestep weighting (see the appendix for details).}
\vspace{-4pt}
\label{tab:dmd_weighting}
    \small
    \setlength{\tabcolsep}{0.5pt}
    \begin{tabularx}{\linewidth}{l|@{\hspace{4pt}}l|cccc|X}
        \toprule
        $\tau$ & $w(\tau)$ & 1NFE & 2NFE & 3NFE & 4NFE & Note \\
        \midrule
        \multirow{8}{*}{Uniform(0,1)} & $(1-\tau)^2/\tau$ & \multicolumn{4}{c|}{Explodes} & Equivalent to the score formulation w/o weighting. \\
        & $(1-\tau)/\tau$ & \multicolumn{4}{c|}{Explodes} &  \\
        & $1-\tau$ & 8.24 & 6.91 & 6.05 & 5.66 & Equivalent to applying the gradient to $\widehat{x}_\tau$. \\
        & $1$ & 4.32 & 3.67 & 3.57 & 3.61 \\
        & $\tau$ & \textbf{3.94} & \textbf{2.99} & \textbf{2.78} & \textbf{2.87} \\
        & $\tau^2$ & 5.31 & 3.45 & 3.13 & 2.99 \\
        & $(1-\tau) \tau$ & 4.32 & 3.45 & 3.44 & 3.43 & DMD weighting without dynamic normalization.\\
        & $(1-\tau) \tau \frac{CS}{\|\hat{x}-x\|_1}$ & 4.24 & 3.24 & 3.22 & 3.22 & DMD weighting. \\
        \midrule
        Beta(2.00, 1) & & \textbf{3.77} & 2.72 & 2.58 & 2.55 & Equivalent to $w(\tau)=\tau$ under uniform sampling.\\
        \rowcolor{gray!20}
        Beta(2.50, 1) & $1$ & 3.85 & \textbf{2.68} & \textbf{2.47} & \textbf{2.46} & \\
        Beta(3.00, 1) & & 4.63 & 2.98 & 2.73 & 2.64 & Equivalent to $w(\tau)=\tau^2$ under uniform sampling.\\
        \bottomrule
    \end{tabularx}

    \end{subtable}
    \par\vspace{8pt}
    \begin{subtable}[t]{0.34\textwidth}
        \vspace{0pt}
    \centering
    \setlength{\tabcolsep}{0.5pt}
    \caption{Conditioning the generator on $r$.}
    \vspace{-4pt}
    \label{tab:generator_conditioning}
    \small
    \begin{tabularx}{\linewidth}{X|cccc}
        \toprule
        Generator & 1NFE & 2NFE & 3NFE & 4NFE \\
        \midrule
        $G(\widehat{x}_0, z)$ & 3.91 & 3.06 & 2.73 & 2.63 \\
        \rowcolor{gray!20}
        $G(\widehat{x}_0, z, r)$ & \textbf{3.85} & \textbf{2.68} & \textbf{2.47} & \textbf{2.46} \\
        \bottomrule
    \end{tabularx}

    \end{subtable}
    \hfill
    \begin{subtable}[t]{0.325\textwidth}
        \vspace{0pt}
    \centering
    \setlength{\tabcolsep}{0.5pt}
    \caption{Adam optimizer betas.}
    \vspace{-4pt}
    \label{tab:adam_betas}
    \small
    \begin{tabularx}{\linewidth}{X|cccc}
        \toprule
        Betas & 1NFE & 2NFE & 3NFE & 4NFE \\
        \midrule
        (0.9, 0.95) & 3.99 & 2.90 & 2.73 & 2.66 \\
        \rowcolor{gray!20}
        (0.0, 0.95) & \textbf{3.85} & \textbf{2.68} & \textbf{2.47} & \textbf{2.46} \\
        \bottomrule
    \end{tabularx}
    

    \end{subtable}
    \hfill
    \begin{subtable}[t]{0.3\textwidth}
        \vspace{0pt}
        \centering
\setlength{\tabcolsep}{0.5pt}
\caption{Optimal transport regularizer.}
\vspace{-4pt}
\label{tab:optimal_transport}
\small
\begin{tabularx}{\linewidth}{X|cccc}
    \toprule
    $\lambda_\mathrm{ot}$ & 1NFE & 2NFE & 3NFE & 4NFE \\
    \midrule
    \rowcolor{gray!20}
    0 & \textbf{3.85} & \textbf{2.68} & \textbf{2.47} & \textbf{2.46} \\
    1e-6 & 3.90 & 2.74 & 2.54 & 2.47 \\
    \bottomrule
\end{tabularx}

    \end{subtable}
\end{table}

\paragraph{Generator conditioning on $r$.}
As shown in Table~\ref{tab:generator_conditioning}, conditioning the generator on the refinement step $r$ consistently improves performance. This conditioning informs the shared generator of its current stage in the refinement process, allowing it to apply a stage-appropriate correction rather than using the same update rule at every step.

\paragraph{Adam optimizer betas.}
As shown in Table~\ref{tab:adam_betas}, we find that setting $\beta_1=0.0$ consistently improves performance. This result suggests that removing first-moment momentum is beneficial for DMD training in our setting. We use this setting for all remaining experiments.

\paragraph{Optimal transport regularizer.}
As shown in Table~\ref{tab:optimal_transport}, the optimal-transport regularizer does not improve performance in our setting. Instead, adding the squared $\ell_2$ regularization term slightly degrades FID relative to the unregularized model. We therefore do not use this regularizer in our final method.

\subsection{Ablation study across different sampling methods}

\begin{table}[t]
    \centering
    \caption{Ablation studies across sampling methods. FID lower is better.}
    \label{tab:sampling_method_ablation}
    \begin{subtable}[t]{\textwidth}
        \centering
\caption{DMD timestep weighting.}
\label{tab:dmd_weighting_all}
\vspace{-4pt}
\small
\setlength{\tabcolsep}{3pt}
\begin{tabularx}{\linewidth}{p{0.16\textwidth}|>{\centering\arraybackslash}p{0.14\textwidth}|cccc|cccc|c}
    \toprule
    & & \multicolumn{4}{c|}{\textbf{Data-space iteration (Ours)}} & \multicolumn{4}{c|}{Flow-renoise} & Flow-map \\
    $\tau$ & $w(\tau)$ & 1NFE & 2NFE & 3NFE & 4NFE & 1NFE & 2NFE & 3NFE & 4NFE & 4NFE \\
    \midrule
    Uniform(0, 1) & $1$ & 4.32 & 3.67 & 3.57 & 3.61 & 5.07 & 4.41 & 4.29 & 4.08 & \\
    Uniform(0, 1) & $\tau$ & 3.94 & 2.99 & 2.78 & 2.87 & 4.64 & 3.82 & 3.47 & 3.42 & \\
    Beta(2.50, 1) & $1$ &  \cellcolor{gray!20}\textbf{3.85} & \cellcolor{gray!20}\textbf{2.68} & \cellcolor{gray!20}\textbf{2.47} & \cellcolor{gray!20}\textbf{2.46} & \cellcolor{gray!20}\textbf{4.45} & \cellcolor{gray!20}\textbf{3.22} & \cellcolor{gray!20}\textbf{2.67} & \cellcolor{gray!20}\textbf{2.75} & \\
    \midrule
    Uniform($t_{r+1}$, 1) & 1 & \textcolor{gray!60}{44.91} & \textcolor{gray!60}{4.08} & \textcolor{gray!60}{2.24} & \textcolor{gray!60}{2.51} & & & & & \cellcolor{gray!20}\textbf{2.55}\\
    Uniform($t_{r+1}$, 1) & $\tau$ & \textcolor{gray!60}{44.20} & \textcolor{gray!60}{4.86} & \textcolor{gray!60}{2.53} & \textcolor{gray!60}{2.49} & & & & & 2.57 \\
    \bottomrule
\end{tabularx}

    \end{subtable}
    \par\vspace{8pt}
    \begin{subtable}[t]{\textwidth}
        \centering
\caption{Classifier-free guidance (CFG).}
\vspace{-4pt}
\label{tab:cfg}
\setlength{\tabcolsep}{3pt}
\small
\begin{tabularx}{\linewidth}{p{0.16\textwidth}|>{\centering\arraybackslash}p{0.14\textwidth}|cccc|cccc|c}
    \toprule
    & & \multicolumn{4}{c|}{\textbf{Data-space iteration (Ours)}} & \multicolumn{4}{c|}{Flow-renoise} & Flow-map \\
    CFG scale & Timestep range & 1NFE & 2NFE & 3NFE & 4NFE & 1NFE & 2NFE & 3NFE & 4NFE & 4NFE \\
    \midrule
    1.0 (disable) & - & 3.85 & 2.68 & 2.47 & 2.46 & 4.45 & 3.22 & 2.67 & 2.75 & \textbf{2.55} \\
    \midrule
    1.1 & \multirow{2}{*}{[0.0, 1.0]} & \textbf{3.06} & 2.61 & 2.60 & 2.59 & 3.63 & 2.80 & 2.74 & 2.97 & 2.96 \\
    1.2 & & 3.28 & 3.23 & 3.26 & 3.27 & \textbf{3.52} & 3.21 & 3.71 & 4.15 & 3.78 \\
    \midrule
    \multirow{3}{*}{1.1} & [0.0, 0.7] & 3.22 & \textbf{2.47} & 2.37 & 2.35 & 3.61 & \textbf{2.70} & 2.55 & 2.74 & 2.93 \\
     & [0.0, 0.6] & \cellcolor{gray!20}3.31 & \cellcolor{gray!20}\textbf{2.47} & \cellcolor{gray!20}\textbf{2.33} & \cellcolor{gray!20}\textbf{2.29} & \cellcolor{gray!20}4.02 & \cellcolor{gray!20}2.89 & \cellcolor{gray!20}\textbf{2.51} & \cellcolor{gray!20}\textbf{2.61} & \cellcolor{gray!20}2.86 \\
     & [0.0, 0.5] & 3.53 & 2.50 & 2.33 & 2.35 & 4.05 & 2.92 & \textbf{2.51} & 2.66 & 2.90 \\
    \bottomrule
\end{tabularx}

    \end{subtable}
    \par\vspace{8pt}
    \begin{subtable}[t]{\textwidth}
        \centering
\caption{Sampling timestep discretization.}
\vspace{-4pt}
\label{tab:dmd_timestep_shift}
\small
\setlength{\tabcolsep}{3pt}
\begin{tabularx}{\linewidth}{p{0.16\textwidth}|>{\centering\arraybackslash}X|>{\centering\arraybackslash}p{0.14\textwidth}|cccc|c}
    \toprule
    & & & \multicolumn{4}{c|}{Flow-renoise} & Flow-map \\
    Shift factor $\gamma$ & Discretization & CFG & 1NFE & 2NFE & 3NFE & 4NFE & 4NFE \\
    \midrule
    0.5 & [1.00, 0.60, 0.33, 0.14] & \multirow{3}{*}{No} & 4.47 & \textbf{2.84} & \textbf{2.67} & 2.84 & 2.52 \\
    1.0 (uniform) & [1.00, 0.75, 0.50, 0.25] & & 4.45 & 3.22 & \textbf{2.67} & 2.75 & 2.55 \\
    2.0 & [1.00, 0.86, 0.67, 0.40] & & \cellcolor{gray!20}\textbf{4.20} & \cellcolor{gray!20}3.44 & \cellcolor{gray!20}2.88 & \cellcolor{gray!20}\textbf{2.72} & \cellcolor{gray!20}\textbf{2.48} \\
    \midrule
    0.5 & [1.00, 0.60, 0.33, 0.14] & \multirow{3}{*}{Yes} & 3.74 & 2.78 & 2.54 & 2.71 & 2.52 \\
    1.0 (uniform) & [1.00, 0.75, 0.50, 0.25] & & \cellcolor{gray!20}4.02 & \cellcolor{gray!20}2.89 & \cellcolor{gray!20}\textbf{2.51} & \cellcolor{gray!20}\textbf{2.61} & 2.86 \\
    2.0 & [1.00, 0.86, 0.67, 0.40] & & \textbf{3.71} & \textbf{2.80} & 2.56 & 2.69 & \cellcolor{gray!20}\textbf{2.37} \\
    \bottomrule
\end{tabularx}

    \end{subtable}
\end{table}

In this section, we conduct ablation studies across data-space iteration, flow-renoise iteration, and flow-map iteration trained using phased DMD~\citep{phaseddmd}. All methods use the same self-rollout training procedure with $R=4$. Flow-renoise produces valid samples at its intermediate steps, allowing us to report FID at every NFE. In contrast, the flow-map model is trained only for four-step generation, so we report its FID only at 4NFE.

\paragraph{DMD weighting across methods.}
Table~\ref{tab:dmd_weighting_all} reports the DMD weighting search for each sampling method. Data-space iteration outperforms the flow-renoise method under every compared weighting setting. Both methods exhibit the same trend: FID improves when the DMD objective places greater emphasis on high-noise timesteps. Because phased DMD trains the mapping from $t_r$ to $t_{r+1}$ by sampling only $\tau \in [t_{r+1}, 1]$, we also train our data-space method using the exact same setting for a matched comparison. Data-space iteration also outperforms the flow-map method under compared weighting setting.

\paragraph{CFG hyperparameter search.}
Table~\ref{tab:cfg} summarizes our classifier-free guidance (CFG)~\citep{cfg} search. Data-space iteration consistently outperforms both flow-renoise and flow-map under all comparable CFG configurations. We first apply standard CFG over the full timestep of $\tau$ and find that it worsens performance. We then restrict CFG to limited ranges of $\tau$~\citep{cfginterval}. Interestingly, CFG degrades the performance of the flow-map method. When enabling CFG, a scale of $1.1$ only over $\tau \in [0.0, 0.6]$ yields the best overall performance across all methods.

\paragraph{Timestep discretization search.} Thus far, both flow-renoise and flow-map have used a uniform trailing flow-time discretization of $[1.0, 0.75, 0.5, 0.25]$ as the baseline~\citep{common}. We use this as our default because it has been adopted by prior few-step generation methods under the same architecture and has yielded state-of-the-art performance on the ImageNet 256$\times$256 setting~\citep{shortcut,alphaflow,tim,tvm,afm}. Following prior work~\citep{sd3}, we experiment with different timestep discretizations obtained by applying the following shift function to the uniform discretization:
\begin{equation}
    \operatorname{shift}(t;\gamma)
    = \frac{\gamma t}{1 + (\gamma-1)t},
    \label{eq:timestep-shift}
\end{equation}
where $\gamma$ is the shift factor. Table~\ref{tab:dmd_timestep_shift} shows that changing the discretization significantly influences the intermediate-step FID of flow-renoise and substantially changes the CFG FID of flow-map. Interestingly, the shift parameter does not exhibit a predictable correlation with final performance across methods or CFG settings, highlighting the difficulty of predicting an optimal timestep discretization.

\subsection{Main comparisons}

\begin{table}[t]
    \centering
    \caption{Main results. FID lower is better.}
    \label{tab:main}
    \small
    \setlength{\tabcolsep}{2.5pt}
    \begin{tabularx}{\linewidth}{lXl|>{\centering\arraybackslash}p{0.15\textwidth}|c|rrrr}
        \toprule
        Method & Paradigm & Discretization & CFG & 250NFE & 1NFE & 2NFE & 3NFE & 4NFE \\
        \midrule
        Flow Matching & flow-ODE & uniform & \multirow{6}{*}{No} & 8.26 \\
        DMD & flow-renoise & uniform & & & 4.45 & 3.22 & 2.67 & 2.75 \\
         & & \textcolor{gray!60}{oracle (best of 3)} & & & \textcolor{gray!60}{4.20} & \textcolor{gray!60}{3.44} & \textcolor{gray!60}{2.88} & \textcolor{gray!60}{2.72} \\
        Phased DMD & flow-map & uniform & & & - & - & - & 2.55 \\
         & & \textcolor{gray!60}{oracle (best of 3)} & & & - & - & - & \textcolor{gray!60}{2.48} \\
        \rowcolor{gray!20}
        \textbf{Data-space DMD} & 
        \textbf{data-space} & \textbf{none} & & & \textbf{3.85} & \textbf{2.68} & \textbf{2.47} & \textbf{2.46} \\
        \midrule
        Flow Matching & flow-ODE & uniform & \multirow{6}{*}{Yes} & 2.06 \\
        DMD & flow-renoise & uniform & & & 4.02 & 2.89 & 2.51 & 2.61 \\
         & & \textcolor{gray!60}{oracle (best of 3)} & & & \textcolor{gray!60}{4.02} & \textcolor{gray!60}{2.89} & \textcolor{gray!60}{2.51} & \textcolor{gray!60}{2.61} \\
        Phased DMD & flow-map & uniform & & & - & - & - & 2.86 \\
         & & \textcolor{gray!60}{oracle (best of 3)} & & & - & - & - & \textcolor{gray!60}{2.37} \\
        \rowcolor{gray!20}
        \textbf{Data-space DMD} & \textbf{data-space} & \textbf{none} & & & \textbf{3.31} & \textbf{2.47} & \textbf{2.33} & \textbf{2.29} \\
        \bottomrule
    \end{tabularx}
\end{table}

Table~\ref{tab:main} reports the main results and addresses two distinct comparisons. Our primary comparison uses the uniform discretization adopted by prior few-step methods. This represents the standard setting in which a practitioner selects a reasonable schedule without training multiple schedule-specific generators. We additionally report the best result from our discretization search as an oracle-style comparison. In the self-rollout setting, obtaining this result requires training a separate generator for every candidate schedule.

Under the standard uniform discretization, data-space iteration consistently outperforms both flow-renoise and flow-map across all reported NFE settings. Even when allowing an oracle selection of the best of three discretizations, flow-map only narrows the gap and remains slightly behind, while flow-renoise continues to lag substantially. Thus, data-space iteration achieves stronger performance without schedule search or schedule-specific training. Additional metrics are provided in the appendix.

Interestingly, in the unguided setting, all DMD-distilled models substantially outperform the flow-matching teacher. We hypothesize that this improvement arises from the mode-seeking behavior of the reverse-KL objective used by DMD, as discussed in prior research~\citep{autoguidance,ddo,cafm}.

\subsection{Visualization}

\begin{figure*}[t]
    \centering
    \setlength{\tabcolsep}{1pt}
    \renewcommand{\arraystretch}{0}
    \newlength{\qualimgwidth}
    \setlength{\qualimgwidth}{\dimexpr(\textwidth-32pt)/9\relax}
    \newcommand{\qualimg}[1]{\includegraphics[width=\qualimgwidth]{#1}}
    \begin{tabular}{@{}cccc@{\hspace{10pt}}cccc@{\hspace{10pt}}c@{}}
        \multicolumn{4}{c}{Data-space iteration (ours)} &
        \multicolumn{4}{c}{Flow-renoise} &
        Flow-map \\[3pt]
        \scriptsize 1NFE & \scriptsize 2NFE & \scriptsize 3NFE & \scriptsize 4NFE &
        \scriptsize 1NFE & \scriptsize 2NFE & \scriptsize 3NFE & \scriptsize 4NFE &
        \scriptsize 4NFE \\[2pt]
        \qualimg{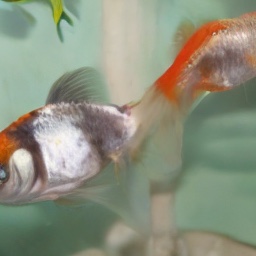} &
        \qualimg{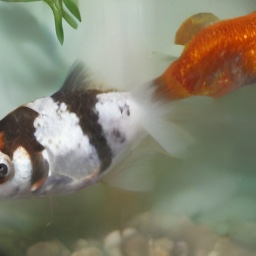} &
        \qualimg{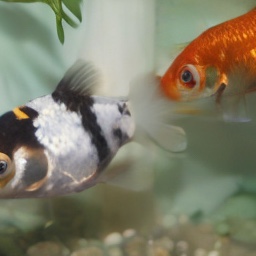} &
        \qualimg{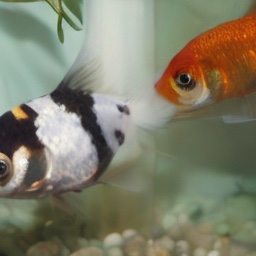} &
        \qualimg{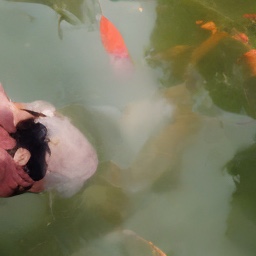} &
        \qualimg{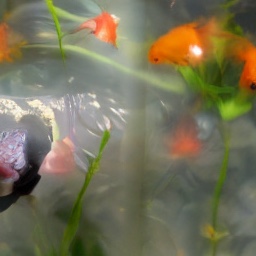} &
        \qualimg{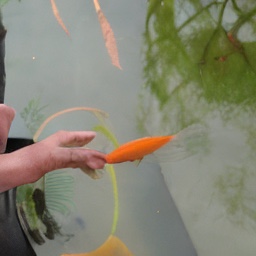} &
        \qualimg{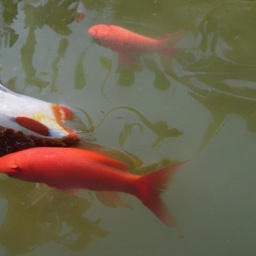} &
        \qualimg{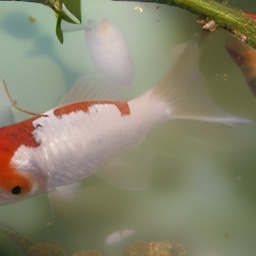} \\[2pt]
        \qualimg{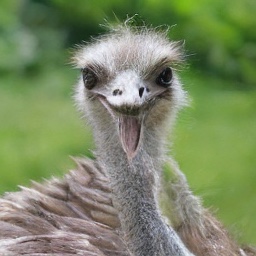} &
        \qualimg{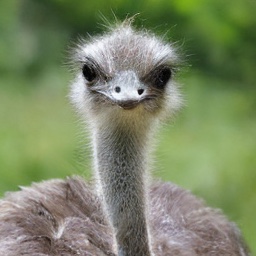} &
        \qualimg{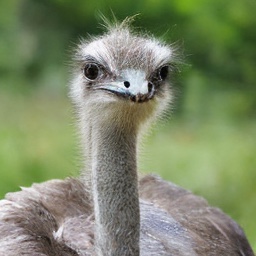} &
        \qualimg{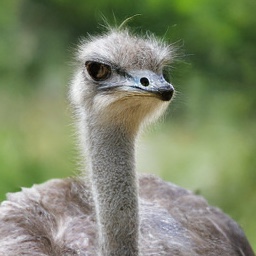} &
        \qualimg{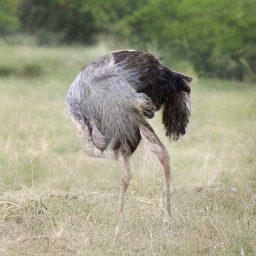} &
        \qualimg{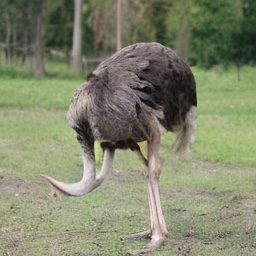} &
        \qualimg{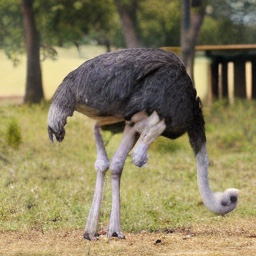} &
        \qualimg{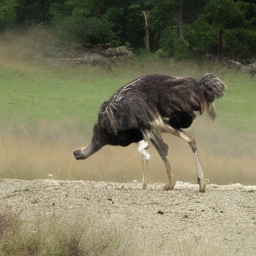} &
        \qualimg{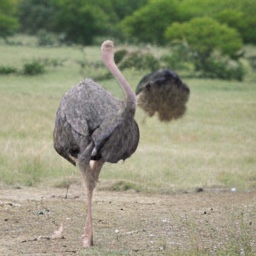} \\[2pt]
        \qualimg{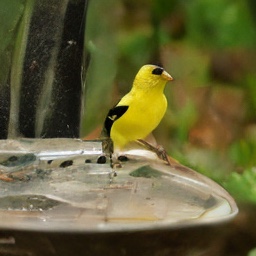} &
        \qualimg{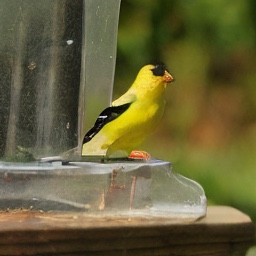} &
        \qualimg{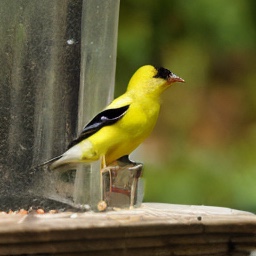} &
        \qualimg{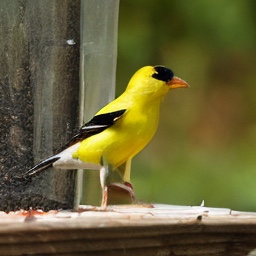} &
        \qualimg{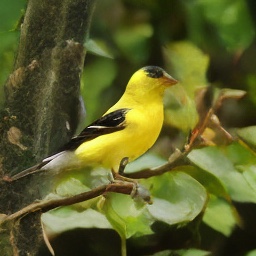} &
        \qualimg{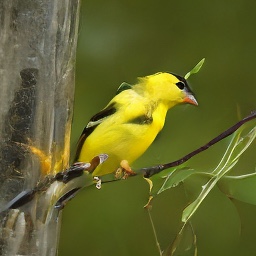} &
        \qualimg{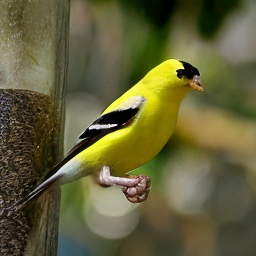} &
        \qualimg{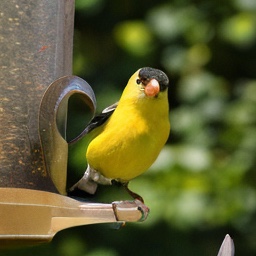} &
        \qualimg{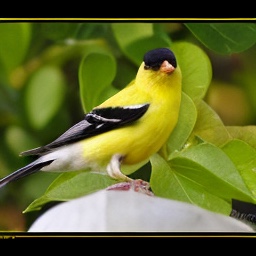} \\[2pt]
        \qualimg{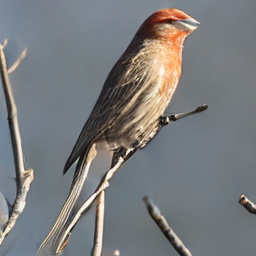} &
        \qualimg{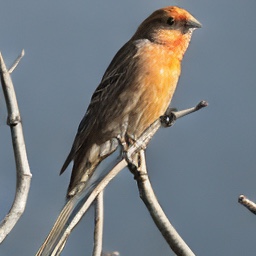} &
        \qualimg{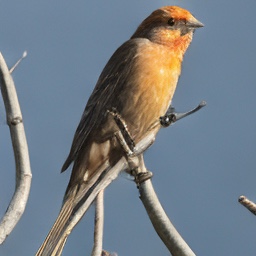} &
        \qualimg{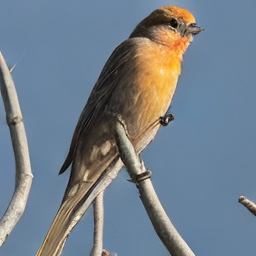} &
        \qualimg{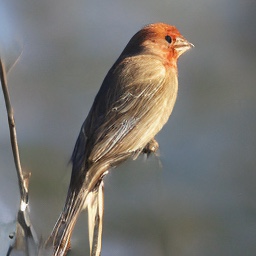} &
        \qualimg{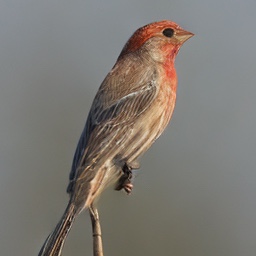} &
        \qualimg{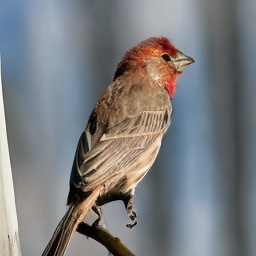} &
        \qualimg{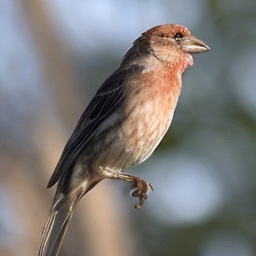} &
        \qualimg{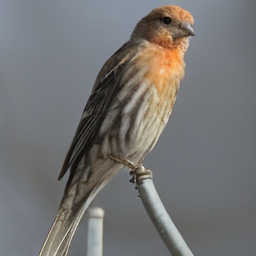}
    \end{tabular}
    \caption{
        Visualization of iterative generation trajectories.
        Each row uses the same initial noise and class condition. Flow-based methods use the uniform discretization.
    }
    \label{fig:qualitative}
\end{figure*}

Figure~\ref{fig:qualitative} visualizes the intermediate generation states of data-space iteration and flow-renoise, together with the final outputs of flow-map.

Despite being trained without the optimal-transport regularizer or any other explicit loss constraint on its refinement trajectory, the data-space iteration generator learns to refine structures and details from its previous prediction. In contrast, flow-renoise exhibits large content changes across iterations: regions that are already satisfactory can be discarded by the renoising process and regenerated with different content, wasting model capacity. Quantitative measurements are provided in the appendix.

\section{Related Work}

\paragraph{Few-step distillation.}
ODE-based distillation methods compress the flow trajectory by matching teacher transitions, velocities, or consistency targets~\citep{progressive,meanflow,shortcut,cm}. However, pointwise regression objectives often produce blurry samples. Recent work shows that distribution-level objectives can substantially improve such distilled generators~\citep{apt,rcm}. Distribution matching distillation (DMD)~\citep{dmd1,dmd2}, building on variational score-distillation~\citep{vsd,diffinstruct}, has become a widely adopted approach. Orthogonal work improves the scaling, parameterization, and score estimation of DMD~\citep{senseflow,twinflow,scoredistillationflowmatching,dataforcing,dmdx}; these advances are complementary to our focus on the generator's iteration space.

\paragraph{Optimizing timestep discretization.}
Prior work has optimized inference-time discretizations for flow models, but uses a single schedule shared across samples~\citep{xue2024accelerating,park2025jump}. Recent methods instead train a separate predictor, conditioned on the same input as the flow model, to produce sample-specific discretizations~\citep{ld3,yuan2026fewstep}. This approach is not directly compatible with self-rollout distillation, where the timestep schedule must be specified during distillation training.

\paragraph{Self-conditioning.}
The closest connection to our method is self-conditioning~\citep{selfcondition,selfcondition2,selfcondition3}. Existing self-conditioning methods use predictions from previous evaluations to refine the model's velocity field, while generation still proceeds by integrating that field along an ODE. Our method instead treats the data prediction itself as the iterative state, so the data-space refinements directly constitute the few-step generation process.

\paragraph{Deep equilibrium models.}
Our formulation also resembles deep equilibrium models; Section~\ref{sec:fixed-point-models} details the key conceptual differences. Recent work applies this paradigm to generative distillation~\citep{deqonestep}, solving for a fixed point in an internal hidden representation before decoding it into an image. This approach requires an equilibrium-specific architecture and solver. In contrast, our method requires minimal architectural changes and is therefore more suitable for distilling existing pretrained generative models.

\paragraph{Looped architectures.}
Looped transformers scale effective depth by repeatedly applying shared transformer blocks~\citep{loopformer,elt}. Recent work has explored this idea for generative modeling by training a looped transformer end to end with a distribution-matching objective~\citep{afm}. However, this approach backpropagates through the full unrolled loop. The resulting computational and memory costs during training can be a hurdle in practice. Data-space iteration is instead a lightweight, drop-in modification to existing distillation training pipelines.

\section{Conclusion}

We introduced data-space iteration for few-step generation. By directly refining data predictions, our method removes the need for a fixed flow-time discretization and integrates into DMD with minimal changes. On class-conditional ImageNet $256\times256$, it outperforms standard uniform-discretization baselines and matches or improves upon schedule-searched variants without requiring schedule-specific training. These results demonstrate that data-space refinement is a simple and effective approach to fast generation.

\paragraph{Limitations.}
Data-space iteration requires self-rollout training. This incurs no additional cost when self-rollout is already required but is a disadvantage otherwise. If ODE-based distillation objectives are preferred over distributional objectives, data-space iteration can be trained with paired data and noise samples, but more efficient training approaches remain to be explored.

\paragraph{Additional contents and future work.}
Following prior few-step generation research, we focus on ImageNet $256\times256$ generation for the main paper. We additionally provide CIFAR-10 results on a UNet architecture in the appendix. We provide the details of hyperparameters, architectures, and code in the appendix. Scaling data-space iteration to text-to-image generation and other modalities is left for future work.

\subsection*{AI use statement}

In this work, we used generative AI tools to assist with drafting and improving the text of the paper, organizing tables and figures, and implementing and double-checking our code. We did not use generative AI tools to run the actual training jobs or curate citations. We have reviewed all AI-assisted work to ensure that the text, code, tables, and figures accurately reflect our intent. We take responsibility for the final content of this work, including text, claims or artifacts produced with the aid of
generative AI.

\subsection*{Acknowledgments}

We sincerely thank Wei Tang for his valuable guidance, insightful discussions, and continued support throughout the development of this project.

\bibliography{iclr2027_conference}
\bibliographystyle{iclr2027_conference}

\clearpage
\appendix
\section{Implementation Details}
\label{app:implementation}

\begin{table}[h]
    \centering
    \caption{Final architecture and training settings used to train the
    data-space iteration generator.}
    \label{tab:appendix_impl}
    \small
    \setlength{\tabcolsep}{8pt}
    \begin{tabularx}{0.72\linewidth}{Xr}
        \toprule
        Configuration & Value \\
        \midrule
        \multicolumn{2}{l}{\textit{Model architecture (DiT-XL/2)}} \\
        Parameters & 674M \\
        FLOPs & 119G \\
        Depth & 28 \\
        Hidden dimension & 1152 \\
        Attention heads & 16 \\
        Patch size & $2\times2$ \\
        \midrule
        \multicolumn{2}{l}{\textit{Data}} \\
        Image shape & $3\times256\times256$ \\
        Latent shape & $4\times32\times32$ \\
        \midrule
        \multicolumn{2}{l}{\textit{Optimization}} \\
        Training epochs & 40 \\
        Global batch size & 256 \\
        Optimizer & Adam \\
        Learning-rate schedule & Constant \\
        Generator learning rate & $1\times10^{-4}$ \\
        Fake-model learning rate & $1\times10^{-4}$ \\
        Adam $(\beta_1,\beta_2)$ & $(0.0, 0.95)$ \\
        Weight decay & $0.0$ \\
        Gradient clipping & No \\
        Fake-model updates per generator update & 4 \\
        DMD timestep distribution & $\tau\sim\operatorname{Beta}(2.5,1)$ \\
        DMD weighting function & $w(\tau)=1$ \\
        EMA decay & $0.9999$ \\
        Numerical precision & TF32 \\
        \midrule
        \multicolumn{2}{l}{\textit{Classifier-free guidance}} \\
        CFG scale & $1.1$ \\
        CFG teacher-timestep range & $[0.0, 0.6]$ \\
        Fake-model warmup & 100 iterations \\
        \bottomrule
    \end{tabularx}
\end{table}

Table~\ref{tab:appendix_impl} summarizes the architecture and training
hyperparameters used in our experiments. We use the XL/2 configuration of the
pretrained SiT backbone for both the generator and the fake flow model. The
generator and fake model are optimized with the same constant learning rate,
and we perform four fake-model updates for every generator update. We do not
use weight decay or gradient clipping. For the DMD objective, the teacher
timestep is sampled as $\tau\sim\operatorname{Beta}(2.5,1)$ and we use the
constant weighting function $w(\tau)=1$. We maintain an exponential moving
average of the generator parameters with a decay of $0.9999$ and use the
averaged parameters for evaluation. All training is performed in TF32
precision.

The added projection for $x_0^{(r)}$ is initialized following the original SiT
initialization and uses no bias. The added $r$ embedding is initialized from
$\mathcal{N}(0, 0.02)$ for both the generator and the fake velocity network.
The fake velocity network receives the refinement index $r$ and is kept
identical across all iteration-method comparisons. During training, we sample
the refinement index simply as $r = i \bmod R$, where $i$ is the current training iteration.

The original SiT-XL/2 model has 675,129,632 parameters. The additional $x_0^{(r)}$ input projection and a $r$ embedding adds 24,192 parameters. However, we can remove the timestep projection MLP and replace it with a constant $t=1$ embedding, which decreases the parameter by 1,626,624. Therefore, our final model has 673,527,200 parameters, which is even less than the original model.

For the classifier-free-guided results, we use a guidance scale of $1.1$ and
apply guidance only over the teacher timestep interval $[0.0, 0.6]$. When training with CFG, we warm up the fake model
for 100 iterations, which prevents training from diverging when CFG is enabled.

\clearpage
\section{Reproducibility}
\label{app:reproducibility}

Results for all models can be reproduced within one day on a GPU server equipped with eight GPUs, each with 80~GB of HBM3 memory, and a high-bandwidth GPU interconnect. Because the final FIDs of data-space iteration and flow-map are close, we run three training runs and use separate evaluation seed. We report the mean FID in the table. The standard deviation is below \(0.05\) for both methods.

\section{Full Metrics}

Table~\ref{tab:appendix_metrics} reports our full evaluation results with and
without classifier-free guidance. Following ADM~\citep{adm}, we use its
provided evaluation code and report spatial FID (sFID), Inception Score (IS),
precision, and recall in addition to FID. We include these additional metrics
for reference, while retaining FID as our primary evaluation metric in
accordance with prior work~\citep{meanflow,imeanflow,shortcut,cm,icm,scm}.
FID and all other metrics are evaluated on 50,000 class-balanced generated
samples.

\begin{table}[h]
    \centering
    \caption{Full metrics.}
    \label{tab:appendix_metrics}
    \begin{subtable}[t]{\linewidth}
        \centering
        \caption{Without guidance.}
        \begin{tabularx}{\linewidth}{X|>{\centering\arraybackslash}p{0.1\textwidth}|ccccc}
            \toprule
            Method & NFE & FID$\downarrow$ & sFID$\downarrow$ & IS$\uparrow$ & Prec.$\uparrow$ & Rec.$\uparrow$ \\
            \midrule
            Flow Matching (SiT)~\citep{sit} & 250 & 8.26 & 6.32 & 131.65 & 0.68 & 0.67 \\
            \midrule
            \multirow{4}{*}{DMD~\citep{dmd1,dmd2}} & 1 & 4.45 & 5.98 & 192.52 & 0.76 & 0.58  \\
            & 2 & 3.22 & 4.67 & 209.88 & 0.79 & 0.57 \\
            & 3 & 2.67 & 4.40 & 223.88 & 0.82 & 0.56 \\
            & 4 & 2.75 & 4.62 & 224.94 & \textbf{0.84} & 0.55 \\
            \midrule
            Phased DMD~\citep{phaseddmd} & 4 & 2.55 & 5.05 & \textbf{245.76} & 0.83 & 0.56 \\
            \midrule
            \multirow{4}{*}{Data-space DMD (Ours)} & 1 & 3.85 & 5.32 & 196.55 & 0.78 & 0.59 \\
            & 2 & 2.68 & 4.59 & 221.69 & 0.81 & 0.58 \\
            & 3 & 2.47 & 4.49 & 228.00 & 0.81 & 0.58 \\
            & 4 & \textbf{2.46} & \textbf{4.48} & 228.59 & 0.81 & \textbf{0.59} \\
            \bottomrule
        \end{tabularx}
    \end{subtable}
    \par\vspace{8pt}
    \begin{subtable}[t]{\linewidth}
        \centering
        \caption{With guidance.}
        \begin{tabularx}{\linewidth}{X|>{\centering\arraybackslash}p{0.1\textwidth}|ccccc}
            \toprule
            Method & NFE & FID$\downarrow$ & sFID$\downarrow$ & IS$\uparrow$ & Prec.$\uparrow$ & Rec.$\uparrow$ \\
            \midrule
            Flow Matching (SiT)~\citep{sit} & 250$\times$2 & 2.06 & 4.49 & 277.50 & 0.83 & 0.59 \\
            \midrule
            \multirow{4}{*}{DMD~\citep{dmd1,dmd2}} & 1 & 4.02 & 6.11 & 209.10 & 0.78 & 0.56 \\
            & 2 & 2.89 & 4.72 & 225.05 & 0.81 & 0.56 \\
            & 3 & 2.51 & 4.40 & 241.14 & 0.84 & 0.55 \\
            & 4 & 2.61 & 4.54 & 243.98 & \textbf{0.85} & 0.54 \\
            \midrule
            Phased DMD~\citep{phaseddmd} & 4 & 2.86 & 5.18 & \textbf{271.99} & \textbf{0.85} & 0.53 \\
            \midrule
            \multirow{4}{*}{Data-space DMD (Ours)} & 1 & 3.31 & 5.19 & 213.21 & 0.79 & 0.58 \\
            & 2 & 2.47 & 4.44 & 238.99 & 0.81 & 0.57 \\
            & 3 & 2.33 & 4.42 & 243.31 & 0.82 & 0.58 \\
            & 4 & \textbf{2.29} & \textbf{4.42} & 242.71 & 0.82 & \textbf{0.58} \\
            \bottomrule
        \end{tabularx}
    \end{subtable}
\end{table}

\clearpage
\section{More Visualizations}
\label{app:diff}

\begin{figure*}[h]
    \centering
    \setlength{\tabcolsep}{1pt}
    \renewcommand{\arraystretch}{0}
    \newlength{\diffimgwidth}
    \setlength{\diffimgwidth}{\dimexpr(\textwidth-22pt)/8\relax}
    \newcommand{\diffimg}[1]{\includegraphics[width=\diffimgwidth]{#1}}
    \begin{tabular}{@{}cccc@{\hspace{10pt}}cccc@{}}
        \multicolumn{4}{c}{Data-space iteration (ours)} &
        \multicolumn{4}{c}{Flow-renoise} \\[3pt]
        \scriptsize 1NFE & \scriptsize 2NFE & \scriptsize 3NFE & \scriptsize 4NFE &
        \scriptsize 1NFE & \scriptsize 2NFE & \scriptsize 3NFE & \scriptsize 4NFE \\[2pt]
        \diffimg{img/qualitative/ours/1_1.jpg} &
        \diffimg{img/qualitative/ours/1_2.jpg} &
        \diffimg{img/qualitative/ours/1_3.jpg} &
        \diffimg{img/qualitative/ours/1_4.jpg} &
        \diffimg{img/qualitative/renoise/1_1.jpg} &
        \diffimg{img/qualitative/renoise/1_2.jpg} &
        \diffimg{img/qualitative/renoise/1_3.jpg} &
        \diffimg{img/qualitative/renoise/1_4.jpg} \\[1pt]
        {} &
        \diffimg{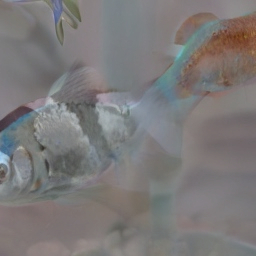} &
        \diffimg{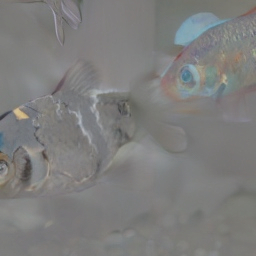} &
        \diffimg{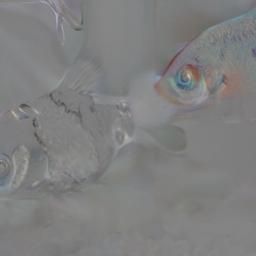} &
        {} &
        \diffimg{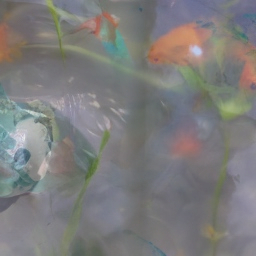} &
        \diffimg{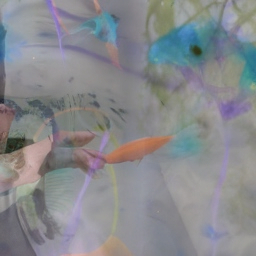} &
        \diffimg{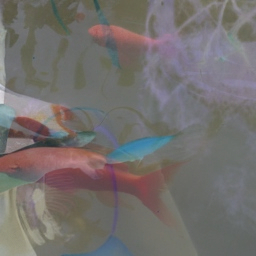} \\[4pt]
        \diffimg{img/qualitative/ours/9_1.jpg} &
        \diffimg{img/qualitative/ours/9_2.jpg} &
        \diffimg{img/qualitative/ours/9_3.jpg} &
        \diffimg{img/qualitative/ours/9_4.jpg} &
        \diffimg{img/qualitative/renoise/9_1.jpg} &
        \diffimg{img/qualitative/renoise/9_2.jpg} &
        \diffimg{img/qualitative/renoise/9_3.jpg} &
        \diffimg{img/qualitative/renoise/9_4.jpg} \\[1pt]
        {} &
        \diffimg{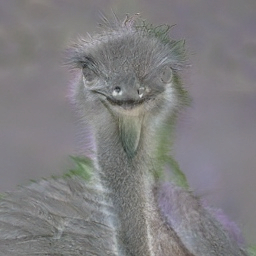} &
        \diffimg{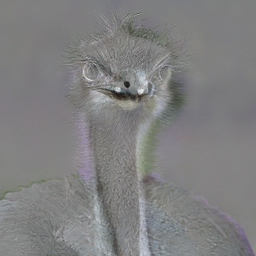} &
        \diffimg{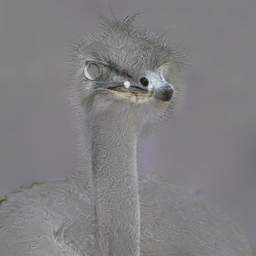} &
        {} &
        \diffimg{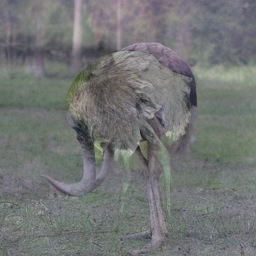} &
        \diffimg{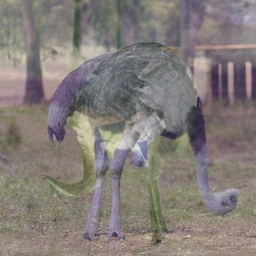} &
        \diffimg{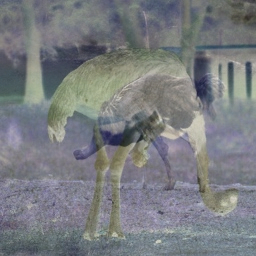} \\[4pt]
        \diffimg{img/qualitative/ours/11_1.jpg} &
        \diffimg{img/qualitative/ours/11_2.jpg} &
        \diffimg{img/qualitative/ours/11_3.jpg} &
        \diffimg{img/qualitative/ours/11_4.jpg} &
        \diffimg{img/qualitative/renoise/11_1.jpg} &
        \diffimg{img/qualitative/renoise/11_2.jpg} &
        \diffimg{img/qualitative/renoise/11_3.jpg} &
        \diffimg{img/qualitative/renoise/11_4.jpg} \\[1pt]
        {} &
        \diffimg{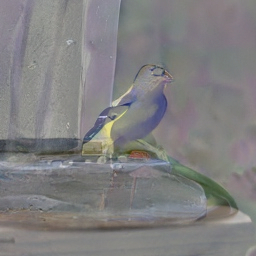} &
        \diffimg{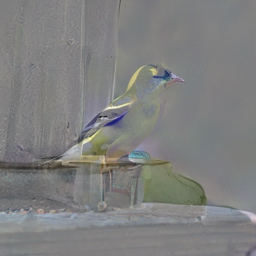} &
        \diffimg{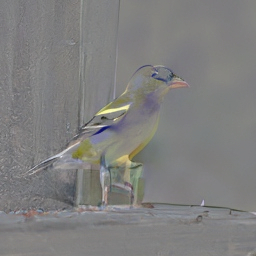} &
        {} &
        \diffimg{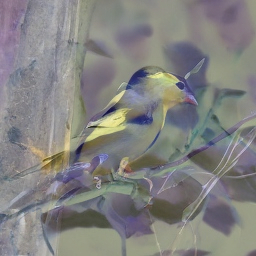} &
        \diffimg{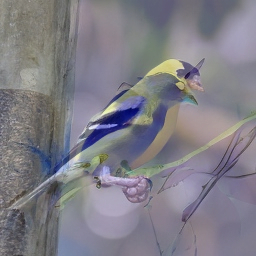} &
        \diffimg{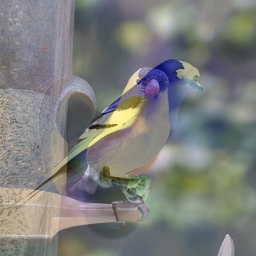} \\[4pt]
        \diffimg{img/qualitative/ours/12_1.jpg} &
        \diffimg{img/qualitative/ours/12_2.jpg} &
        \diffimg{img/qualitative/ours/12_3.jpg} &
        \diffimg{img/qualitative/ours/12_4.jpg} &
        \diffimg{img/qualitative/renoise/12_1.jpg} &
        \diffimg{img/qualitative/renoise/12_2.jpg} &
        \diffimg{img/qualitative/renoise/12_3.jpg} &
        \diffimg{img/qualitative/renoise/12_4.jpg} \\[1pt]
        {} &
        \diffimg{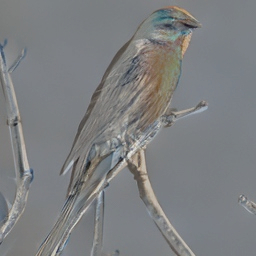} &
        \diffimg{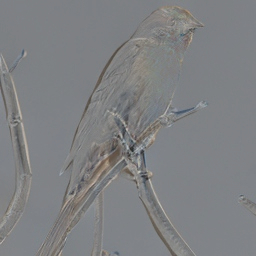} &
        \diffimg{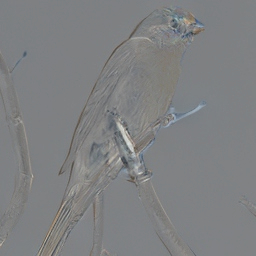} &
        {} &
        \diffimg{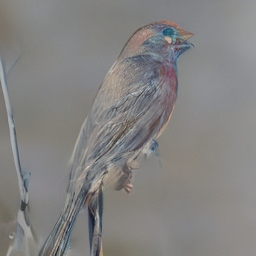} &
        \diffimg{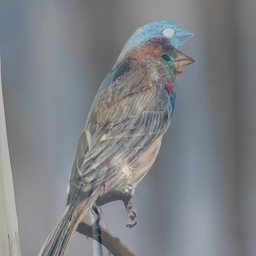} &
        \diffimg{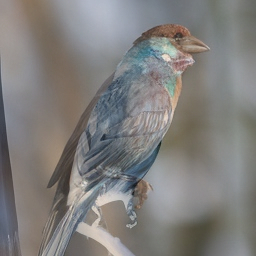}
    \end{tabular}
    \caption{
        Visualization of iterative generation trajectories and their stepwise differences.
        For each sample, the first row shows the generated images and the second row shows
        the signed difference between consecutive steps, aligned beneath the later step.
        Zero difference is mapped to mid-gray using the same fixed normalization for all images.
    }
    \label{fig:diff}
\end{figure*}

Figure~\ref{fig:diff} visualizes the differences between consecutive refinement steps.

Table~\ref{tab:appendix_trajectory} quantifies the change between consecutive rollout steps. Specifically, for each step $r$, we compute the average L2 difference $\mathbb{E}\lVert x_0^{(r)} - x_0^{(r+1)} \rVert_2$ over 50k class-balanced generated samples. The difference is measured after VAE decoding in RGB pixel space, with pixel values normalized to $[0,1]$.

\begin{table}[h]
    \centering
    \caption{Average L2 difference between rollout steps measured over 50k samples.}
    \label{tab:appendix_trajectory}
    \begin{tabular}{l|c|c|c}
        \toprule
        Method & $r=1$ & $r=2$ & $r=3$ \\
        \midrule
        Flow-renoise & 82.72 & 51.18 & 29.91 \\
        Data-space & 59.65 & 44.25 & 35.98 \\
        \bottomrule
    \end{tabular}
\end{table}

\clearpage
\section{Code}
\label{app:code}

Algorithm~\ref{alg:dmd_code} gives the core PyTorch implementation. The two rollout functions expose the only difference between data-space iteration and flow-renoise iteration; \texttt{dmd\_step} implements their shared DMD objective. The sampled index $r\in\{0,\ldots,R-1\}$ identifies the refinement step being optimized: steps $0,\ldots,r-1$ are evaluated without gradient tracking, followed by the gradient-tracked step $r$ that produces $\hat{x}_0^{(r+1)}$. Here, \texttt{data} contains a latent batch and its class labels, while \texttt{gen\_step} selects the alternating generator or fake-model update. Distributed training, optimizer steps, and EMA updates are omitted for clarity.

\begin{algorithm}[h]
\caption{Core DMD training step with data-space and flow-renoise rollouts.}
\label{alg:dmd_code}
\begin{lstlisting}[
    language=Python,
    basicstyle=\ttfamily\scriptsize,
    keywordstyle=\color{blue!70!black}\bfseries,
    commentstyle=\color{green!45!black}\itshape,
    stringstyle=\color{red!65!black},
    emph={torch,gen,v_real,v_fake,rollout_fn},
    emphstyle=\color{teal!70!black},
    columns=fullflexible,
    keepspaces=true,
    showstringspaces=false
]
def rollout_data_space(gen, z, c, r, track_grad):
    x_hat = torch.zeros_like(z)
    for i in range(r + 1):
        with torch.set_grad_enabled(track_grad and i == r):
            x_hat = z - gen(x_hat, z, c, i)
    return x_hat


def rollout_flow_renoise(gen, z, c, r, track_grad, R=4):
    x_hat = torch.zeros_like(z)
    for i, t_r in enumerate(torch.arange(R, 0, -1, device=z.device)[:r + 1] / R):
        t_r = t_r.expand(z.shape[0])
        t_r4 = t_r[:, None, None, None]
        eps = z if i == 0 else torch.randn_like(z)
        x_tr = (1.0 - t_r4) * x_hat + t_r4 * eps
        with torch.set_grad_enabled(track_grad and i == r):
            x_hat = x_tr - t_r4 * gen(x_tr, t_r, c)
    return x_hat


def dmd_step(gen, v_real, v_fake, data, r, rollout_fn, gen_step=True, use_cfg=False):
    x, c = data
    B = x.shape[0]
    z, eps = torch.randn_like(x), torch.randn_like(x)
    tau = torch.distributions.Beta(2.5, 1.0).sample((B,)).to(x.device)
    tau4 = tau[:, None, None, None]

    x_hat = rollout_fn(gen, z, c, r, track_grad=gen_step)
    x_tau = (1.0 - tau4) * x_hat + tau4 * eps

    if not gen_step:
        return F.mse_loss(v_fake(x_tau, tau, c, r), eps - x_hat)

    with torch.no_grad():
        v_pos = v_real(x_tau, tau, c)
        if use_cfg:
            c_null = torch.full_like(c, 1000)
            v_null = v_real(x_tau, tau, c_null)
            v_cfg = v_null + 1.1 * (v_pos - v_null)
            v_pos = torch.where(tau4 <= 0.6, v_cfg, v_pos)
        direction = v_pos - v_fake(x_tau, tau, c, r)

    return (x_hat * direction).mean()
\end{lstlisting}
\end{algorithm}

\clearpage
Algorithm~\ref{alg:phased_dmd_code} shows the corresponding Phased DMD baseline. After rolling out to the endpoint $t_{r+1}$ of phase $r$, it samples a DMD noising time $\tau$ from the valid interval of that phase and re-noises the intermediate flow state before applying the DMD objective.

\begin{algorithm}[h]
\caption{Core Phased DMD training step.}
\label{alg:phased_dmd_code}
\begin{lstlisting}[
    language=Python,
    basicstyle=\ttfamily\scriptsize,
    keywordstyle=\color{blue!70!black}\bfseries,
    commentstyle=\color{green!45!black}\itshape,
    stringstyle=\color{red!65!black},
    emph={torch,gen,v_real,v_fake},
    emphstyle=\color{teal!70!black},
    columns=fullflexible,
    keepspaces=true,
    showstringspaces=false
]
def rollout_flow_map(gen, z, c, r, track_grad, R=4):
    x_tr = z
    dt = 1.0 / R
    for i in range(r + 1):
        t_r = torch.full((z.shape[0],), 1.0 - i * dt, device=z.device)
        with torch.set_grad_enabled(track_grad and i == r):
            x_tr = x_tr - dt * gen(x_tr, t_r, c)
    return x_tr


def phased_dmd_step(gen, v_real, v_fake, data, r, gen_step=True, use_cfg=False, R=4):
    x, c = data
    B = x.shape[0]
    z, eps = torch.randn_like(x), torch.randn_like(x)

    x_next = rollout_flow_map(gen, z, c, r, track_grad=gen_step, R=R)
    t_next = torch.full((B,), 1.0 - (r + 1) / R, device=x.device)
    u = 0.02 + 0.98 * torch.rand(B, device=x.device) # Uniform(0.02, 1)
    tau = t_next + (1.0 - t_next) * u
    t_next4, tau4 = t_next[:, None, None, None], tau[:, None, None, None]

    a = (1.0 - tau4) / (1.0 - t_next4)
    b = (tau4**2 - (t_next4 * a)**2)**0.5
    x_tau = a * x_next + b * eps

    if not gen_step:
        target = (
            ((1.0 - t_next4)**2 * tau4 + (1.0 - tau4) * t_next4**2)
            / ((1.0 - t_next4)**2 * b) * eps
            - x_next / (1.0 - t_next4)
        )
        return F.mse_loss(v_fake(x_tau, tau, c, r), target)

    with torch.no_grad():
        v_pos = v_real(x_tau, tau, c)
        if use_cfg:
            c_null = torch.full_like(c, 1000)
            v_null = v_real(x_tau, tau, c_null)
            v_cfg = v_null + 1.1 * (v_pos - v_null)
            v_pos = torch.where(tau4 <= 0.6, v_cfg, v_pos)
        direction = v_pos - v_fake(x_tau, tau, c, r)

    return (x_next * direction).mean()
\end{lstlisting}
\end{algorithm}

\clearpage
\section{DMD Weighting Functions and Their Connections}
\label{app:dmd-weighting}

\paragraph{$\boldsymbol{w(\tau)=(1-\tau)^2/\tau}$.}
The original DMD generator update can be written as the stop-gradient
surrogate
\begin{equation}
    \mathcal{L}_{\mathrm{score}}
    =-\mathbb{E}\!\left[
        \operatorname{sg}\!\left[
            s_{\mathrm{real}}(x_\tau,\tau)-s_{\mathrm{fake}}(x_\tau,\tau)
        \right]^{\!\top}x_\tau
    \right],
\end{equation}
where $s_{\mathrm{real}}$ and $s_{\mathrm{fake}}$ denote the real and fake
scores, respectively. For the linear path
$x_\tau=(1-\tau)x_0+\tau x_1$, with $x_1$ distributed as standard Gaussian noise, a
velocity predictor and its corresponding score satisfy
\begin{equation}
    s(x_\tau,\tau)
    =-\frac{\mathbb{E}[x_1\mid x_\tau]}{\tau}
    =-\frac{x_\tau}{\tau}-\frac{1-\tau}{\tau}v(x_\tau,\tau),
\end{equation}
because $\mathbb{E}[x_1\mid x_\tau]=x_\tau+(1-\tau)v(x_\tau,\tau)$. The term
$-x_\tau/\tau$ cancels between the real and fake models, and hence
\begin{equation}
    -\left(s_{\mathrm{real}}(x_\tau,\tau)-s_{\mathrm{fake}}(x_\tau,\tau)\right)^{\!\top}x_\tau
    =\frac{1-\tau}{\tau}
    \left(v_{\mathrm{real}}(x_\tau,\tau)-v_{\mathrm{fake}}(x_\tau,\tau)\right)^{\!\top}x_\tau.
\end{equation}
Converting the score difference to a velocity difference therefore contributes
a factor of $(1-\tau)/\tau$.  When this update is applied to the clean generator
prediction $\widehat{x}_0$ through the re-noising operation, there is an
additional factor of $1-\tau$, since
\begin{equation}
    x_\tau=(1-\tau)\widehat{x}_0+\tau\varepsilon,
    \qquad
    \frac{\partial x_\tau}{\partial \widehat{x}_0}=1-\tau.
\end{equation}
More precisely,
\begin{align}
    \nabla_\theta \mathcal{L}_{\mathrm{score}}
    &=-\mathbb{E}\!\left[
        (1-\tau)\operatorname{sg}\!\left[
            s_{\mathrm{real}}(x_\tau,\tau)-s_{\mathrm{fake}}(x_\tau,\tau)
        \right]^{\!\top}
        \frac{\partial \widehat{x}_0}{\partial\theta}
    \right] \\
    &=\mathbb{E}\!\left[
        \frac{(1-\tau)^2}{\tau}\operatorname{sg}\!\left[
            v_{\mathrm{real}}(x_\tau,\tau)-v_{\mathrm{fake}}(x_\tau,\tau)
        \right]^{\!\top}
        \frac{\partial \widehat{x}_0}{\partial\theta}
    \right].
\end{align}
Therefore, an unweighted score-based DMD update backpropagated through $x_\tau$
is equivalent to using $w(\tau)=(1-\tau)^2/\tau$ in our velocity-based objective on
$\widehat{x}_0$.

\paragraph{$\boldsymbol{w(\tau)=(1-\tau)/\tau}$.}
Alternatively, consider applying the original score difference directly to
$\widehat{x}_0$, with the stop-gradient surrogate
\begin{equation}
    \mathcal{L}_{\mathrm{score},x_0}
    =-\mathbb{E}\!\left[
        \operatorname{sg}\!\left[
            s_{\mathrm{real}}(x_\tau,\tau)-s_{\mathrm{fake}}(x_\tau,\tau)
        \right]^{\!\top}\widehat{x}_0
    \right].
\end{equation}
In this case, there is no additional Jacobian factor from re-noising. Using
the score--velocity relation above gives
\begin{equation}
    \nabla_\theta \mathcal{L}_{\mathrm{score},x_0}
    =\mathbb{E}\!\left[
        \frac{1-\tau}{\tau}\operatorname{sg}\!\left[
            v_{\mathrm{real}}(x_\tau,\tau)-v_{\mathrm{fake}}(x_\tau,\tau)
        \right]^{\!\top}
        \frac{\partial \widehat{x}_0}{\partial\theta}
    \right].
\end{equation}
Thus, $w(\tau)=(1-\tau)/\tau$ is equivalent to the original score-difference
formulation applied directly to $\widehat{x}_0$. Both weightings diverge as
$\tau\to0$. Following the original DMD setup~\citep{dmd1}, we therefore restrict
the sampled timesteps to $\tau\geq \tau_{\min}=0.02$ for these ablations;
nevertheless, both runs still diverge in our experiments.

\paragraph{$\boldsymbol{w(\tau)=1-\tau}$.}
This weighting is equivalent to applying the velocity difference directly to
$\widehat{x}_\tau$, since backpropagating through
$\widehat{x}_\tau=(1-\tau)\widehat{x}_0+\tau\varepsilon$ contributes the factor $1-\tau$.

\paragraph{$\boldsymbol{w(\tau)=1}$.}
This weighting is equivalent to applying the velocity difference directly to
$\widehat{x}_0$.

\paragraph{$\boldsymbol{w(\tau)=\tau}$ and $\boldsymbol{w(\tau)=\tau^2}$.}
These weightings have no direct theoretical connection.  We use them to place
increasing emphasis on high-noise timesteps, which empirically improves FID.

\paragraph{$\boldsymbol{w(\tau)=(1-\tau) \tau}$.}
The original DMD update~\citep{dmd1} uses the adaptive weight
\begin{equation}
    \widetilde{w}_\tau
    =\frac{\sigma_\tau^2}{\alpha_\tau}
    \frac{CS}{\lVert\mu_{\mathrm{base}}(x_\tau,\tau)-\widehat{x}_0\rVert_1}
\end{equation}
on $\alpha_\tau(s_{\mathrm{fake}}-s_{\mathrm{real}})$.  Since
\begin{equation}
    s_{\mathrm{fake}}-s_{\mathrm{real}}
    =\frac{\alpha_\tau}{\sigma_\tau^2}
    (\mu_{\mathrm{fake}}-\mu_{\mathrm{real}}),
\end{equation}
the noise-schedule factors reduce to $\alpha_\tau$ on the denoised-prediction
difference.  For the linear path, $\alpha_\tau=1-\tau$, $\sigma_\tau=\tau$, and
$\mu=x_\tau-\tau v$, so
\begin{equation}
    \mu_{\mathrm{fake}}-\mu_{\mathrm{real}}
    =\tau(v_{\mathrm{real}}-v_{\mathrm{fake}}).
\end{equation}
Thus, after omitting the adaptive normalization, the effective velocity
weighting is $\alpha_\tau \tau=(1-\tau)\cdot \tau$.

\paragraph{$\boldsymbol{w(\tau)=(1-\tau) \tau
\frac{CS}{\lVert\widehat{x}-x\rVert_1}}$.}
This retains the adaptive gradient normalization of the original DMD update,
where $C$ is the number of channels and $S$ is the number of spatial
locations.

\paragraph{Beta distributions.}
Table~\ref{tab:beta-distributions} shows the three probability density
functions. Since $\operatorname{Beta}(a,1)$ has density
$p(\tau)=a\tau^{a-1}$, $\operatorname{Beta}(2.00,1)$ is equivalent to
$w(\tau)=\tau$ and $\operatorname{Beta}(3.00,1)$ to $w(\tau)=\tau^2$, each up
to a constant scaling factor; $\operatorname{Beta}(2.50,1)$ interpolates
between them.

\begin{table}[h]
    \centering
    \setlength{\tabcolsep}{2pt}
    \begin{tabular}{@{}ccc@{}}
        $\operatorname{Beta}(2.00,1)$ &
        $\operatorname{Beta}(2.50,1)$ &
        $\operatorname{Beta}(3.00,1)$ \\
        \includegraphics[width=0.32\textwidth]{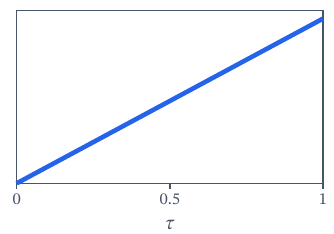} &
        \includegraphics[width=0.32\textwidth]{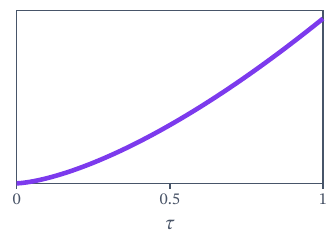} &
        \includegraphics[width=0.32\textwidth]{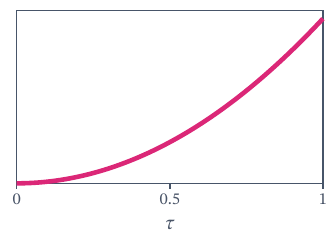}
    \end{tabular}
    \caption{Probability density functions of the three Beta distributions.}
    \label{tab:beta-distributions}
\end{table}

\section{CIFAR-10 Experiment}

Following the convention of prior work~\citep{meanflow}, we additionally
conduct experiments on unconditional CIFAR-10 generation~\citep{cifar10} (32$\times$32)
using the SongUNet architecture provided by EDM~\citep{edm} (55M). For simplicity,
we do not conduct additional hyperparameter ablations or use auxiliary
techniques such as data augmentation. We first train the teacher using the
standard flow-matching objective with a linear schedule. The resulting
flow-matching teacher achieves an FID-50k of 3.54 at 250 NFE. We then use the same
experimental setup as in the main text to train DMD, phased DMD, and our
proposed data-space DMD. The resulting FIDs are reported in
Table~\ref{tab:appendix_cifar10}.

\begin{table}[h]
    \centering
    \caption{CIFAR-10 results.}
    \label{tab:appendix_cifar10}
    \begin{tabular}{l|cccc}
        \toprule
        Method & 1NFE & 2NFE & 3NFE & 4NFE \\
        \midrule
        DMD & \textbf{11.21} & 11.14 & 9.93 & 10.03 \\
        Phased DMD & - & - & - & 11.72 \\
        Data-space DMD & 12.20 & \textbf{8.28} & \textbf{8.32} & \textbf{8.62} \\
        \bottomrule
    \end{tabular}
\end{table}

For one-step generation, data-space iteration and flow-renoise reduce to
essentially the same formulation; therefore, we do not expect either method to
have an inherent advantage over the other. At 1 NFE, data-space iteration
outperforms flow-renoise on ImageNet, as reported in the main text, but performs
slightly worse on CIFAR-10. We attribute this discrepancy to differences in
optimization dynamics. Our analysis therefore focuses primarily on performance
at NFE $>1$, where we attribute the observed differences to the distinct
iteration strategies.

\end{document}